\documentclass[journal]{IEEEtran}

\usepackage{cite}
\usepackage{graphicx}
\usepackage{amsmath}
\usepackage{amsfonts}
\usepackage{algorithmic}
\usepackage{array}
\usepackage[caption=false,font=footnotesize]{subfig}
\usepackage{url}
\usepackage{multirow}

\begin{document}

\title{Mixture Model Framework for Traumatic Brain Injury Prognosis Using Heterogeneous Clinical and Outcome Data}

\author{Alan~D.~Kaplan,~\IEEEmembership{Member,~IEEE,}
        Qi~Cheng,~\IEEEmembership{Senior Member,~IEEE,}
        K.~Aditya~Mohan,~\IEEEmembership{Senior Member,~IEEE,}
        Lindsay~D.~Nelson,
        Sonia~Jain,
        Harvey~Levin,
        Abel~Torres-Espin,
        Austin~Chou,
        J. Russell~Huie,
        Adam~R.~Ferguson,
        Michael~McCrea,
        Joseph~Giacino,
        Shivshankar Sundaram,
        Amy J. Markowitz,
        and Geoffrey~T.~Manley
\thanks{A. D. Kaplan, Q. Cheng, K. A. Mohan, and S. Sundaram are with the Lawrence Livermore National Laboratory, Livermore, CA, 94550 USA e-mail: kaplan7@llnl.gov.}
\thanks{L. Nelson and M. McCrea are with the Medical College of Wisconsin.}
\thanks{S. Jain is with the University of California, San Diego.}
\thanks{H. Levin is with the Baylor College of Medicine.}
\thanks{J. Giacino is with the Massachusetts General Hospital and Harvard Medical School.}
\thanks{A. Torres-Espin, A. Chou, J. R. Huie, A. R. Ferguson, A. J. Markowitz, and G. T. Manley are with the University of California, San Francisco.}
\thanks{Manuscript received [Month] [Day], 2020; revised [Month] [Day], 2020.}}

\markboth{}
{Shell \MakeLowercase{\textit{et al.}}: Mixture Model Framework for Traumatic Brain Injury Prognosis Using Heterogeneous Clinical and Outcome Data}

\maketitle

\begin{abstract}
Prognoses of Traumatic Brain Injury (TBI) outcomes are neither easily nor accurately determined from clinical indicators.
This is due in part to the heterogeneity of damage inflicted to the brain, ultimately resulting in diverse and complex outcomes. 
Using a data-driven approach on many distinct data elements may be necessary to describe this large set of outcomes and thereby robustly depict the nuanced differences among TBI patients’ recovery. 
In this work, we develop a method for modeling large heterogeneous data types relevant to TBI.
Our approach is geared toward the probabilistic representation of mixed continuous and discrete variables with missing values.
The model is trained on a dataset encompassing a variety of data types, including demographics, blood-based biomarkers, and imaging findings.
In addition, it includes a set of clinical outcome assessments at 3, 6, and 12 months post-injury.
The model is used to stratify patients into distinct groups in an unsupervised learning setting.
We use the model to infer outcomes using input data, and show that the collection of input data reduces uncertainty of outcomes over a baseline approach.
In addition, we quantify the performance of a likelihood scoring technique that can be used to self-evaluate the extrapolation risk of prognosis on unseen patients.

\end{abstract}


\IEEEpeerreviewmaketitle

\section{Introduction}

\IEEEPARstart{T}{raumatic} Brain Injury (TBI) is a leading cause of death and disability in the United States.
In 2014 over 2.8 million emergency department visits and hospitalizations were attributed to TBI in the United States, resulting in over 56 thousand deaths.\footnote{https://www.cdc.gov/traumaticbraininjury/data/tbi-edhd.html}
Internationally, more than 50 million people have a TBI each year \cite{Maas2017-yb, Dewan2018-vh}.

Despite these large numbers, there are no targeted treatments for TBI, and only limited systematic understanding of prognosis, owing in part to the immense complexity of the human brain and the diverse and heterogeneous presentation of injury.
The injuries themselves vary greatly in severity, location of affected areas within the brain, and physical damage imparted to the tissue.
Further, many factors may influence outcome and recovery, including prior medical history, genetics, and demographics.
Effects of the injury can evolve over many months following the initial insult, resulting in varying levels of measurable clinical sequelae as across domains of physical, psychological, and/or cognitive deficits \cite{Millis2001-wh, Rabinowitz2014-hc}.
Ultimately, many factors may combine to affect the type and severity of a complex outcome, including the type and intensity of intervention \cite{Dikmen2017-xe, Mollayeva2019-gl}.

From a precision medicine perspective, no two patients are exactly the same, and this is readily apparent in the study of TBI.
The challenge, therefore, is to model prognosis from available individualized data in order to design effective interventions in a precision medicine framework.
An important step in this direction is the utilization of available collections of TBI data to uncover associations between demographics, preinjury medical history, clinical, imaging findings, blood-based biomarkers, and outcome assessments.
Finding associations across these domains, if carefully validated, could assist in making accurate prognoses of recovery for individual TBI patients. 
Using this approach to leverage all available data will inform understanding of the natural history of TBI, and in the acute and sub-acute clinical setting as an aid in decision making for early clinical management, and targeting of treatment interventions.

The \underline{T}ransforming \underline{R}esearch and \underline{C}linical \underline{K}nowledge in \underline{T}raumatic \underline{B}rain \underline{I}njury (TRACK-TBI) study has advanced a data-driven view of TBI through the longitudinal collection of diverse clinical, imaging, and outcome metrics from patients who sustained a TBI and presented to one of the participating level 1 trauma centers for care\footnote{https://tracktbi.ucsf.edu/transforming-research-and-clinical-knowledge-tbi}.
The data collection utilizes the National Institute of Neurological Disorders and Stroke (NINDS) Common Data Elements (CDEs) for TBI, which are recommended in order to facilitate comparison across studies \cite{Maas2010-qc, Yue2013-gz}.
The TRACK-TBI Pilot dataset (TRACK-Pilot) is comprised of multimodal data from 586 participants with varying severities of acute TBI. 
Collected data domains included demographics, previous medical history, injury details, clinical information, blood-based biomarker measures, head Computed Tomography (CT) annotations, Magnetic Resonance Imaging (MRI) annotations, and outcome measures at 3, 6 and 12 months post-injury.

Here, we describe a methodology for modeling the TRACK-Pilot dataset to reveal latent structure in the data.
In addition, we evaluate prediction performance on outcome measures and measure the reduction in uncertainty of the outcomes relative to chance and a baseline measure.
Our work aims to discover associations between 
injury and outcome, which could result in improved prognostic models and clinical decision making tools.
Our methods are designed to model the collection of this large quantity of information jointly with the broad array of outcome metrics.
Given the large number of missing values, we explicitly incorporate missingness into the model.
This enables the inclusion of all subjects during training, and the ability to perform inference using any subset of the data elements.
In addition, we use the input test likelihood 
to evaluate whether an unseen subject (not trained on) is within the distribution that the model was trained on.

\subsection{Relation to Existing Work}
Prognostic models for TBI which predict unfavorable TBI outcomes at 6 months as defined by the Glasgow Outcome Scale (GOS) score have been developed which utilize logistic regression \cite{Steyerberg2008-tv}.
In this approach, 26 measures including demographics, vital signs, Glasgow Coma Scale (GCS) score, and CT were used to predict outcome.
This approach has been validated on several other large datasets with slight modification on the set of predictor variables \cite{Roozenbeek2012-fg}.
There is evidence to suggest that such regression modeling approaches for outcome prognosis which utilize a curated set of predictor variables are limited in their clinical utility \cite{Silverberg2015-xo}.
Our approach extends this methodology beyond regression and utilizes a much larger spectrum of predictor measures.

Imaging modalities have been studied for binary classification of TBI using control subjects, including connectivity metrics derived from diffusion MRI using regularized multivariate Gaussian models \cite{Shaker2015-yo} and Generalized Linear Models \cite{Mitra2016-vt}.
A Support Vector Machine approach using a combination of diffusion MRI with resting state functional MRI (fMRI) was developed for TBI prediction  \cite{Vergara2017-aw}.
A method based on logistic regression has been applied to predict the GOS using 17 predictor variables which include head CT annotations and blood measures \cite{Taslimitehrani2014-cv}.
In our work, we leverage expert annotations of both CT and clinical Magnetic Resonance Imaging (MRI), totaling 135 imaging variables.
These image-based measures are combined with 367 non-imaging variables and modeled jointly with 352 outcome variables.

Within the TRACK-TBI study datasets, several data-driven approaches have been leveraged to study subsets of the data.
Topological Data Analysis was used to stratify patients based on a total of 17 selected features, which included CT-derived measures and outcome metrics \cite{Nielson2017-nq}.
Multiple variable groups constructed from CT, MRI, diffusion MRI derived measures, and outcomes were used in a correlative study \cite{Yuh2014-wk}.
Combined proteomics and CT measures were used to predict Glasgow Outcome Scale-Extended (GOSE) using a Principal Components Analysis (PCA) approach \cite{Huie2019-lq}.
The risk of Post-Traumatic Stress Disorder (PTSD) and depression was analyzed using statistical tests on a larger cohort than included in the TRACK-Pilot dataset \cite{Stein2019-gi}.
The present work differs in that we construct a modeling framework which encompasses all 854 variables from the TRACK-Pilot study jointly, instead of targeted subsets of the dataset.
In addition, we 
evaluate the use of the model
to provide
a measure of similarity between a test subject and the subjects used for training.

Heterogeneous data are defined as containing data elements of different types \cite{Wang2017-iz}.
Since many modeling and prediction techniques do not explicitly allow for modeling heterogeneous data, some form of data transformation is often applied \cite{Wang2017-iz}.
Kernel approaches are appealing in this regime, since they involve an initial step of computing a distance matrix across samples, resulting in uniform pairwise measures \cite{Pavlidis2001-ds, Lewis2006-ay, Xiang2018-ld}.
Learning custom kernels, or distance metrics is an active area of research \cite{Luo2018-mx}.
Exploratory factor analysis approaches represent the data in terms of latent parameters, often in linear combinations (see, e.g. \cite{McLachlan2004-mh}).
An extension of the Mixture of Factor Analyzers (MFA) to accommodate missing values was developed that incorporates imputation within the iterative training algorithm \cite{Wei2020-qd}.
Unsupervised machine learning methods which are probabilistic in nature are often designed to explicitly ingest heterogeneous data.
A graphical model using multinomial emissions was developed for collections of patient data in electronic medical records \cite{Pivovarov2015-zu}.
A mixture of product compositions including 7 static and binned summary statistics of 6 time series for sepsis modeling uses a combination of categorical, Gaussian, gamma, and exponential distributions \cite{Mayhew2018-pt}.
In our work, we do not perform any transformations of the data prior to modeling.
Our methods expand on these probabilistic methods in terms of the number of data elements used, and the explicit modeling of missingness.

\subsection{Organization}
The rest of the paper is organized as follows.
In Section \ref{sec_methods}, we present the data and describe methodological development.
Results for modeling, prediction, and extrapolation risk evaluation are presented in Section \ref{sec_results}.
Discussion and Conclusions are presented in Sections \ref{sec_discussion} and \ref{sec_conclusion}, respectively.

\section{Data and Methods} \label{sec_methods}
    \subsection{Data Collection and Description}
    The TRACK-Pilot dataset was used for the experiments in this paper \cite{Yue2013-gz}.
    This data collection effort aimed to construct a comprehensive picture of TBI from multiple modalities and a rich series of longitudinal outcome assessments.
    These modalities include scalar measurements, CT imaging, and 3T MRI structural imaging.
    Variables were selected to be compliant with the NINDS CDEs for TBI.
    All data collection was approved by the IRB for the participating sites (University of California, San Francisco/Zuckerberg San Francisco General Hospital, San Francisco, CA; University of Pittsburgh Medical Center, Pittsburgh, PA; and University Medical Center Brackenridge, Austin, TX).\footnote{Clinical trial titled ``Transforming Research and Clinical Knowledge in TBI Pilot," clinicaltrials.gov identifier: NCT01565551}
    
    A total of 586 subjects are included in the study.
    The variables in the dataset fall under these broad categories: Demographics, Previous Medical History, Injury, Clinical, Blood Specimen, Imaging, and Outcomes \cite{Yue2013-gz}.
    \emph{Demographics} includes age, sex, and socio-economic measures.
    \emph{Previous Medical History} includes basic medical status including cardiac, gastrointestinal, and neurological.
    \emph{Injury} variables assess the nature of the injury, including the type (car accident, fall, etc.) and safety equipment used (helmet, seat belt, etc.).
    \emph{Clinical} data include information collected at the time of hospital presentation, such as vital signs, GCS score (a measure of level of consciousness following a TBI), and basic laboratory blood tests.
    \emph{Blood Specimen} tests include concentrations of proteins elevated in TBI populations such as Glial Fibrillary Acidic Protein (GFAP) concentration, and inflammatory and degenerative disease indicators.
    \emph{Imaging} refers to radiologist assessments derived from brain imaging (CT, MRI) which include normal/abnormal classification, bone fracture, and hemorrhage evaluation.
    For a complete list and description, see the TRACK-TBI website\footnote{https://tracktbi.ucsf.edu/researchers}.
    We refer to this collection of non-outcome data as inputs.
    
    The clinical outcome assessment battery, or \emph{Outcomes}, includes
    patient-reported (standardized questionnaires), observer-reported (structured interviews), and performance-based (standardized cognitive tests) measures, designed to characterize the complex landscape of TBI outcomes.
    They are broadly categorized under these domains: Neuropsychological Impairment, Psychological Status, TBI-Related Symptoms, Perceived Health-Related Quality of Life, and Physical Function \cite{Nelson2017-hc}.
    Each of these domains is assessed via one or more tests, each of which in turn include multiple variables.
    All of these tests were performed in person at 6 months post-injury, whereas a subset was performed over the phone at 3 and 12 months post injury.
    Included in these assessments is the GOSE, a common interview-based, examiner-rated  measure of global function/injury-related disability that was collected at 3, 6, and 12 months post-injury. 
    We refer to outcome variables at 3, 6, or 12 months post-injury with \textsf{-\{3,6,12\}M} appended to the variable name, e.g. \textsf{GOSE-12M}.

    In this work, we develop models to capture joint dependencies between all variables in TRACK-Pilot, including outcome measures recorded at 3, 6, and 12 months post-injury.
    Several variables were removed that did not have any variability (i.e., were either always missing or the same value).
    The total number of variables used was 854, which comprise 502 input and 352 outcome variables.
    Table \ref{tbl_count} shows the number of variables within each category.
    
    \begin{table}
    \renewcommand{\arraystretch}{1.3}
    \caption{Variable Count by Category}
    \label{tbl_count}
    \centering
    \begin{tabular}{l||c}
    \textbf{Variable Category} & \textbf{Count}\\
    \hline
    Demographics &   18\\
    Previous Medical History	&   85\\
    Injury  &   22\\
    Clinical &   122\\
    Blood Specimen	&   120\\
    Imaging  &   135\\
    Outcomes & 352\\
    \hline
    \emph{Total} & 854\\
    \end{tabular}
    \end{table}
    
    We provide a description of several outcome measures for which results are presented in this paper:
    
    \begin{itemize}
        \item \emph{Glasgow Outcome Scale-Extended}
        The \textsf{GOSE} is an interview-based assessment of injury-related disability on a 1-8 point scale: 1-Dead, 2-Persistent Vegetative State, 3-Lower Severely Disabled, 4-Upper Severely Disabled, 5-Lower Moderately Disabled, 6-Upper Moderately Disabled, 7-Lower Good Recovery, and 8-Upper Good Recovery \cite{Wilson1998-hs, Teasdale1998-qp}.
        \item \emph{Neurological Assessment}
        The \textsf{Neuro} variable encodes how differently a subject is acting as compared to typical behavior on a 1-6 point scale: 1-Normal, 6-Very different.
        \item \emph{Return to work}
        \textsf{Return} is an important measure is working status.
        The categorical values for this measure are: Not returned, Sheltered, Partial, Full, N/A, and Unknown.
        \item \emph{Post Traumatic Stress Disorder Checklist-Civilian Version}
        The \textsf{PCL} variable is a 17 item scale with a maximum of 5 points per item, resulting in a 17-85 point scale, with higher values indicating more severe PTSD symptoms \cite{Weathers1991-jt}.
        \item \emph{Satisfaction With Life Scale}
        The \textsf{SWLS} variable is a score in the range of 5-35, with lower values indicating poor general life satisfaction and higher values indicating high satisfaction \cite{Diener1985-cu}.
        \item \emph{Rivermead Post-Concussion Symptoms Questionnaire}
        The \textsf{RPQ} variable is a 0-52 point scale measuring responses to 13 items, each on a 0-4 scale.
        This test measures common TBI symptoms including headache, dizziness, cognitive difficulties, irritability, and depression which are often sequelae of relatively mild TBI, i.e. “concussion” \cite{King1995-uy}.
    \end{itemize}
    
    In addition, we describe several input variables that are presented in this paper:
    
    \begin{itemize}
        \item \emph{Age}
        The \textsf{Age} of the patient in years.
        \item \emph{Glasgow Coma Scale}
        The \textsf{GCS} is a measure of depth and duration of coma and impaired consciousness following TBI based on the ability to follow instructions related to eye opening, motor, and verbal responses \cite{Teasdale1974-uu, Sternbach2000-xh}.
        The scale, which is widely reported as the sum of the 3 component scores,  is a commonly used index of  brain injury severity. Sum scores range from 3-15, with lower values indicating lower levels of consciousness. The GCS  is administered upon admission to the emergency department.
        \item \emph{CT-Marshall}
        The \textsf{Marshall} score is an index of brain pathology identified by the  head CT which is typically obtained within 24 hours post-injury \cite{Marshall1992-pc}.
        This score ranges from 1 indicating no visible acute trauma-related pathology in the brain, to 6 indicating the presence of large trauma-related lesions.
        \item \emph{Ubiquitin carboxyl-terminal esterase L1}
        The \textsf{UCH-L1} variable is a highly brain-specific enzyme that can be measured in serum, and is responsible for disposing of unneeded proteins in the brain \cite{Day2010-ii}. It is released in higher concentrations in the setting of TBI.
    \end{itemize}

    \subsection{Modeling Heterogeneous Data with Missing Values} \label{sec_modeling}
        In this section we define the base mixture model (Section \ref{sec_mixture_model}), its components (Section \ref{sec_var_dists}), and extend it to handle missing values (Section \ref{sec_missing}).
        \subsubsection{Mixture model} \label{sec_mixture_model}
        We construct a joint probability distribution over the collection of both input and outcome variables using a latent variable approach.
        This distribution outputs the likelihood of any combination of variables and its parameters will be estimated using training data.
        The collection of observed variables variables for a single subject is denoted as 
        $\boldsymbol{x}=\left[x_1, x_2, \ldots, x_V\right]$.
        First, we model each variable $x_v$ with a distribution $f\left(x_v ; \boldsymbol{\theta}_v\right)$, where $\boldsymbol{\theta}_v$ is a vector of parameters for variable $x_v$.
        The specific form of the distribution for each variable is chosen from a family of distributions depending on their support (see Table \ref{table_dists}).
        We then link the variables using a 
        mixture model
        with distribution:
        \begin{equation*}
            f\left(\boldsymbol{x}\right) = 
            \sum_{z=1}^{k} \alpha_z \prod_{v=1}^p 
            f\left(x_v ; \boldsymbol{\theta}_{z, v}\right),
        \end{equation*}
        where each variable indexed by $v$ has one set of parameters for each 
        component $z$ of the mixture model: $\boldsymbol{\theta}_{z, v}$.
        The mixing coefficients are constrained by $\alpha_z > 0$ and $\sum_z \alpha_z = 1$.
        The number of 
        components $k$ is 
        referred to as the model order, and
        the number of variables is $p$.
        Under our model, we assume that the variables are independent given a mixture component.
        The product term, therefore, combines all of the single variable distributions. 
        The Expectation Maximization (EM) procedure is used to estimate the parameters (Section \ref{sec_EM}).
        Using this framework, we are able to control the complexity of the model (Section \ref{sec_model_sel}), while capturing dependence between the high dimensional set of variables.
        Given a trained model we can perform inference of outcome variables by computing conditional distributions of the outcomes given the input variables (Section \ref{sec_infer}).

\subsubsection{Variable distributions} \label{sec_var_dists}
        
        The atomic units of this model are distributions chosen on the basis of the variable type and  domain.
        Table \ref{table_dists} shows the selection criteria that we used to choose these variable distributions.
        The Type can be either Continuous, Ordinal, or Categorical.
        Ordinal and Categorical variables are discrete valued, where Ordinal variables have an ordering, and Categorical variables do not.
        The Domain is the set of permissible values for each variable.
        Some continuous variables cover both positive and negative numbers (Real), whereas others are Nonnegative.
        Gaussian distributions are used for real-valued variables, but are not well suited to strictly nonnegative values since the support of the Gaussian is the entire real line.
        For nonnegative valued variables, we use a Zero-Inflated Gamma distribution, which is a Gamma distribution amended with a probability for a value of 0.
        This distribution takes the form:
        \begin{equation*}
            f\left(x_v\right) = 
            \begin{cases}
            t_{z, v}, & \text{if } x_v=0 \\
            \left(1 - t_{z, v}\right)
            \frac{1}{\Gamma(k)\theta^k}x^{k-1}e^{\frac{-x_v}{\theta}}, & \text{else,}
            \end{cases}
        \end{equation*}
        where $\theta>0 $ and $k>0$ are the Gamma distribution parameters, and $\Gamma$ is the Gamma function.
        For Ordinal variables, we choose to use a quantized Gaussian distribution.
        This gives additional flexibility over the more natural Poisson dsitribution, since we can model scale and location, whereas with the Poisson distribution, the scale increases with the location.
        The quantized Gaussian distribution is  a Categorical distribution over discrete points $x_v \in \mathcal{D}$ with the shape constrained by the Gaussian function:
        $$
        f\left(x_v\right) = \frac{\phi\left(x_v; \mu_v, \sigma_v^2\right)}{\sum_{x\in\mathcal{D}}\phi\left(x; \mu_v, \sigma_v^2\right)},
        $$
        where
        $$
        \phi\left(x;\mu_v, \sigma_v^2\right) =  \frac{1}{2\pi\sigma_v^2}e^{-\frac{(x_v - \mu_v)^2}{\sigma_v^2}}.
        $$
        
        The mixture model formulation incorporates mixtures of these single variable distributions, resulting in a multidimensional characterization of the joint distribution that is more expressive than their unimodal versions.
        
        \begin{table}[tpb]
            \caption{Variable Distribution Selection}
            \label{table_dists}
            \begin{center}
            \begin{tabular}{l|l|l}
            \hline
            \multirow{2}{*}{Type}        & \multirow{2}{*}{Domain}            & Distribution \\ & &  (Parameters)\\
            \hline
            \multirow{2}{*}{Continuous}  & \multirow{2}{*}{Real}              & Gaussian      \\ & & ($\mu, \sigma^2$)\\
            \multirow{2}{*}{Continuous}  & \multirow{2}{*}{Nonnegative Real}     & Inflated Gamma      \\ & &  ($t, \theta, k$)\\
            \multirow{2}{*}{Ordinal}     & \multirow{2}{*}{Positive Integer} & Quantized Gaussian \\ & & ($\mu, \sigma^2$)\\
            \multirow{2}{*}{Categorical} & \multirow{2}{*}{Symbolic}          & Categorical   \\ & &  ($\boldsymbol{p}$)\\
            \hline
            \end{tabular}
            \end{center}
        \end{table}

        \subsubsection{Missing values} \label{sec_missing}
\begin{figure}[!t]
        \centering
        \includegraphics[width=2.5in]{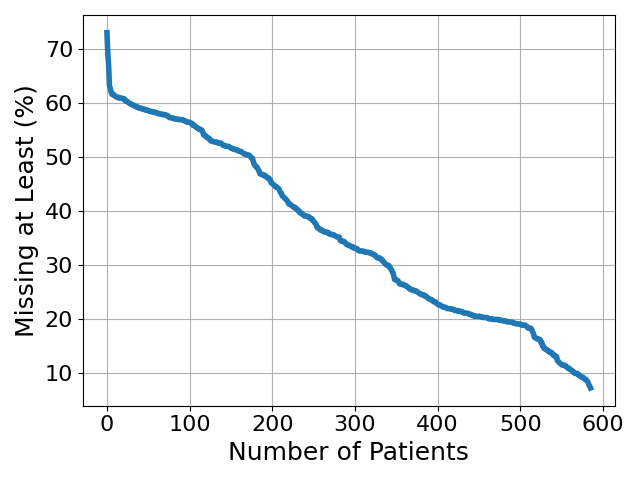}
        \caption{Number of patients (abscissa) with at least a given proportion of missing values (ordinate).}
        \label{fig_missing}
        \end{figure}

        As is common in clinical datasets, TRACK-Pilot contains values that are missing for a variety of reasons (e.g. the subject may not be well enough to complete a test, or has voluntarily decided not to complete a test).
        In this work, we treat all of these missingness causes equally.
        Fig. \ref{fig_missing} shows the number of patients that are missing at least a given percentage of their records.

        Modeling data which contain missing elements is a topic of rigorous statistical analysis (see e.g. \cite{Little2019-nn}).
        In order to fully leverage every available subject record in the data, we incorporate missingess explicitly into our model.
        We accomplish this by encoding the presence of missing values with  Bernoulli random variables $Q_{z,v}$, with probability of missingness $q_{z,v}$, for each latent component and variable combination.

        The overall model has the form,
        \begin{equation} \label{eq_model}
            f\left(\boldsymbol{x}\right)  = \sum_{z=1}^{k} \alpha_z 
            \prod_{v=1}^p 
            \begin{cases}
            q_{z, v}, & \text{if missing} \\
            \left(1 - q_{z, v}\right)f\left(x_v ; \boldsymbol{\theta}_{z, v}\right), & \text{else}
            \end{cases}
        \end{equation}
        where $q_{z, v}$ is the missingness probability of variable $v$ in  component $z$.

        \subsection{Estimation using EM} \label{sec_EM}
        In this section we describe the estimation of model parameters defined by the model in Section \ref{sec_modeling}. This includes variable distribution parameters (Section \ref{sec_est_model_params}) and the number of components (Section \ref{sec_model_sel}).
        
        \subsubsection{Estimating model parameters} \label{sec_est_model_params}
        Given the number of components $k$, the remaining parameters are estimated using an EM procedure.
        Using a collection of training data 
        $X=\left[\boldsymbol{x}_1, \boldsymbol{x}_2, \ldots, \boldsymbol{x}_n\right]$
        , where $\boldsymbol{x}_s$ is the training data for subject $s$, we seek to estimate the following parameters: 
        $\alpha_z$, $\boldsymbol{\theta}_{z, v}$, and $q_{z, v}$ for $z \in \{1, \ldots, k\}$ and $v \in \{1, \ldots, p\}$.
        We follow the standard iterative EM procedure that fits neatly into our framework (see e.g. \cite{Moon1996-kb}).
        
        We introduce an indicator variable, $Z$ with realization $z$ which assigns
        one of the mixture components to each subjects' data $\boldsymbol{x}_s$ 
        (see e.g. \cite{Dempster1977-gh}).
        The EM iterations alternate between the E-Step and M-Step.
        First, we initialize all of the parameters randomly. Then we iterate between the following steps \cite{Dempster1977-gh, Moon1996-kb}:
        \begin{itemize}
            \item E-Step: Compute the expectation of the complete-data log-likelihood, $Q\left( \boldsymbol{\theta} | \boldsymbol{\theta}^{prev}\right) = \mathbb{E}[\log f\left(\boldsymbol{x}, z; \boldsymbol{\theta}\right) | \boldsymbol{x}, \boldsymbol{\theta}^{prev}]$. 
            In this expression, the expectation is taken with respect to the  conditional distribution of $Z$ given data $\boldsymbol{x}$ and the previous set of parameters $\boldsymbol{\theta}^{prev}$.
            \item M-Step: Maximize this expectation with respect to the parameters: $\boldsymbol{\theta} = \arg \max_{\boldsymbol{\theta}} Q\left( \boldsymbol{\theta} | \boldsymbol{\theta}^{prev}\right)$.
        \end{itemize}
        This process is continued until convergence. Within the E-Step, we need to compute the likelihood of $Z$ using the current parameter estimates $w_{s, z}=\Pr\left[Z=z | \boldsymbol{x}_s; \boldsymbol{\theta}\right]$. These values appear in the Q function, which can be expanded as,
        \begin{equation*}
            \begin{split}
            & Q\left( \boldsymbol{\theta} | \boldsymbol{\theta}^{prev}\right) = \\
            & \sum_{s=1}^n \sum_{z=1}^k w_{s, z} \left( \log \alpha_{s, z} + \sum_{v=1}^p \log f\left(x_{s, v}; q_{s, v}, \boldsymbol{\theta}_{s, v}\right) \right).
            \end{split}
        \end{equation*}
        
        
        Since in this expression contributions from each variable do not interact with each other, we can optimize each variable separately.
        Here we describe the M-step for the less frequently used inflated Gamma and quantized Gaussian distributions.
        We drop the $v$ subscript as the same expressions are used for each variable of the same type.
        
        The parameter estimates given in this subsection are contained within the parameter vector $\boldsymbol{\theta}$ of our model (\ref{eq_model}). 
        For the inflated Gamma distributions, we employ the commonly used Newton's method to compute iterative estimates of the parameters (see e.g. \cite{Choi1969-ax}) using,
        $$
        t_z = \frac{\sum_s w_{s,z}I_{x_s = 0}}{\sum_s w_{s,z}},
        $$
        $$
        \theta_z = \frac{\sum_s w_{s,z} x_s}{k\sum_s w_{s,z}},
        $$
        $$
        k_z \leftarrow k_z - \frac{\ln k_z - \psi\left(k\right) - \gamma_z}{\frac{1}{k} - \psi'\left(k\right)},
        $$
        where
        $$
        \gamma_z = \ln \frac{\sum_s w_{s,z} x_s}{\sum_s w_{s,z}} - \frac{\sum_s w_{s,z}\ln x_s}{\sum_s w_{s,z}}.
        $$
        
        The indicator function $I_{(\cdot)}$ is equal to 1 when the condition $(\cdot)$ is true, and 0 otherwise.
        In the iterative estimates for $k_z$, the digamma function is denoted $\psi\left(k\right)$.
        
        For the quantized Gaussian, we compute parameter estimates during the M-step as,
        $$
        \mu_z = \frac{\sum_s w_{s,z}x_s}{\sum_s w_{s,z}},
        $$
        $$
        \sigma_z^2 = \frac{\sum_s w_{s,z}\left(x_s - \mu_z\right)^2}{\sum_s w_{s,z}}.
        $$
        
        In addition to the parameters associated with variable distributions, we also estimate the missingness probabilities, $q_z$.
        In order to do this, we accumulate sufficient statistics of the missing token and update parameters within the EM framework using,
        $$
        q_{z} = \frac{\sum_s w_{s,z} I_{x_s=\text{missing}}}{\sum_s w_{s,z}}.
        $$
        
        \subsubsection{Model selection using a linear search} \label{sec_model_sel}
        
        The model order, or number of components, $k$ is estimated using a linear search over the Bayesian Information Criterion (BIC) for each model \cite{Hastie2009-ba}.
        The BIC score is computed as
        $$
        \text{BIC}\left(k\right)=\frac{1}{2}T_k \ln N - \ln f_k\left(\boldsymbol{x}\right),
        $$
        where $T_k$ is the total parameter count of the model with order $k$, $N$ is the sample size, and $\ln f_k\left(\boldsymbol{x}\right)$ is the log-likelihood of the model with order $k$.
        The model order corresponding to the lowest value of the BIC is then selected.
        
    \subsection{Performing Inference and Extrapolation Risk Evaluation} 
        Once the model is trained we can compute conditional distributions in order to do inference of outcomes.
        We derive the inference equations in Section \ref{sec_infer}, performance evaluation metrics in Section \ref{sec_perf}, and baseline methods in Section \ref{sec_chancebase}.
        
        \subsubsection{Inferring distributions of unknown variables} \label{sec_infer}
        
        The inference procedure operates on the set of inputs, $\boldsymbol{x}_{\textsf{in}}$, and the set of outcomes $\boldsymbol{x}_{\textsf{out}}$.
        We would like to compute the likelihood of all outcomes simultaneously given the inputs, $f\left(\boldsymbol{x}_{\textsf{out}} | \boldsymbol{x}_{\textsf{in}}\right)$ using a trained model of all of the variables, $f\left(\boldsymbol{x}\right) = f\left(\boldsymbol{x}_{\textsf{in}}, \boldsymbol{x}_{\textsf{out}}\right)$.
        In general, we can define any subsets of variables $\boldsymbol{x}_{\textsf{in}}$ and $\boldsymbol{x}_{\textsf{out}}$ to perform this inference on.
        This quantity can be expressed as:
        \begin{equation} \label{eq_inf}
            f\left(\boldsymbol{x}_{\textsf{out}} | \boldsymbol{x}_{\textsf{in}}\right) = 
            \sum_{z=1}^{k} \Pr\left[Z=z|\boldsymbol{x}_{\textsf{in}}\right]f\left(\boldsymbol{x}_{\textsf{out}};\boldsymbol{\theta}_z\right).
        \end{equation}
        The first term inside the summation is the posterior probability of the indicator $Z$ given the input (known) data $\boldsymbol{x}_{\textsf{in}}$, and the second term is the likelihood of the data to be inferred $\boldsymbol{x}_{\textsf{out}}$ conditioned on the latent variable.
        This can be viewed as a two step approach, where first the known data are used to project into a latent space with coordinates defined by probabilities of the latent components.
        Secondly, distributions over the unknown variables are computed based on this projection using a linear combination of component-wise distributions.
        
        Equation (\ref{eq_inf}) is derived by noting that
        \begin{equation*}
        \begin{split}
            f\left(\boldsymbol{x}_{\textsf{out}} | \boldsymbol{x}_{\textsf{in}}\right) & = \frac{1}{f\left(\boldsymbol{x}_{\textsf{in}}\right)} \sum_{z=1}^{k} \Pr\left[Z=z\right]f\left(\boldsymbol{x}_{\textsf{in}}, \boldsymbol{x}_{\textsf{out}};\boldsymbol{\theta}_z\right)
            \\
            & =
            \frac{1}{f\left(\boldsymbol{x}_{\textsf{in}}\right)} \sum_{z=1}^{k} \Pr\left[Z=z\right]f\left(\boldsymbol{x}_{\textsf{in}};\boldsymbol{\theta}_z\right) f\left(\boldsymbol{x}_{\textsf{out}};\boldsymbol{\theta}_z\right)
            \\
            & =
            \sum_{z=1}^{k} \Pr\left[Z=z|\boldsymbol{x}_{\textsf{in}}\right] f\left(\boldsymbol{x}_{\textsf{out}};\boldsymbol{\theta}_z\right),
        \end{split}
        \end{equation*}
        where we make use of the fact that $f\left(\boldsymbol{x}_{\textsf{in}}, \boldsymbol{x}_{\textsf{out}};\boldsymbol{\theta}_z\right) = f\left(\boldsymbol{x}_{\textsf{in}};\boldsymbol{\theta}_z\right) f\left(\boldsymbol{x}_{\textsf{out}};\boldsymbol{\theta}_z\right)$ and $f\left(\boldsymbol{x}_{\textsf{in}};\boldsymbol{\theta}_z\right) = \Pr\left[Z=z|\boldsymbol{x}_{\textsf{in}}\right]f\left(\boldsymbol{x}_{\textsf{in}}\right)/\Pr\left[Z=z\right]$.
        
        With respect to missing values in $\boldsymbol{x}_{\textsf{in}}$, we consider two options for inference.
        In the first, we evaluate the conditional likelihood in (\ref{eq_inf}) by evaluating the missingness probability for variables that are missing, directly affecting the posterior computation $\Pr[Z=z|\boldsymbol{x}_{\textsf{in}}]$.
        The implicit assumption under this scenario is that the missingness is informative for inference of the outcome.
        
        In the second version, we ignore missing values during inference.
        This is accomplished by using the following likelihood function during inference, which marginalizes out the missing variable,
        \begin{equation}
        \begin{split}
            f\left(\boldsymbol{x}\right)  & = \sum_{z=1}^{k} \Pr\left[Z=z\right] \cdot
            \\
            &
            \prod_{v=1}^V 
            \begin{cases}
            1, & \text{if missing} \\
            f\left(x_v ; \boldsymbol{\theta}_{z, v}\right), & \text{else.}
            \end{cases}
        \end{split}
        \end{equation}
        This effectively skips over variables which have missing values.
        This scenario represents the case where missingness is assumed to not be informative.
        Note that there is no issue with differing numbers of non-missing values across subjects for inference in this approach, since we are computing the posterior distribution independently for each subject.
        
        \subsubsection{Prediction and performance evaluation} \label{sec_perf}
        To predict an outcome, we use the inference procedure described in Section \ref{sec_infer} to compute the distribution of the outcome variable given a trained model.
        The distribution of this variable can be used in various ways depending on the application: select the most likely outcome, rank the likelihoods of outcome values, or report the entire distribution.
        In the case of ordinal-valued outcome variables, a measure of distance is not easily defined.
        We chose to use absolute error since we are able to more readily interpret it than the squared error, which exaggerates larger errors.
        To evaluate prediction performance given the inferred distribution of an outcome variable $x$,  $f\left(x\right)=f\left(x|\boldsymbol{x}_\textsf{in}\right)$ and the true value $x_{true}$, we compute the Expected Absolute Error (EAE) for Continuous and Ordinal variables,
        \begin{equation}\label{eq_EAE}
            \sum_x f\left(x\right)|x - x_{true}|.
        \end{equation}
        This is an expectation with respect to the distribution inferred by the model: $\mathbb{E}|X - x_{true}|$.
        For Categorical variables, we compute the probability of error
        $$
            1 - f\left(x_{true}\right).
        $$
        The EAE can be computed for every outcome variable and averaged across patients.
        
        For validation, we use a leave-one-out scheme.
        For each left out subject, models are trained on the remaining subjects, covering a range of model orders.
        Then we infer distributions for the outcome variables using the input variables (Section \ref{sec_infer}) and compute the EAE.
        
        \subsubsection{Baseline and chance performance} \label{sec_chancebase}
        We compare our performance to those of a chance classifier which outputs the uniform distribution, and a baseline model.
        For ordinal variables, the EAE for the chance classifier is empirically computed for each subject by substituting the uniform distribution into (\ref{eq_EAE}) and obtaining,
        $$
            \frac{1}{|\mathcal{X}|}\sum_{x \in \mathcal{X}} |x - x_{true}|,
        $$
        where $|\mathcal{X}|$ is the cardinality of the variable.
        This quantity will vary across subjects for the same variable, as $X_{true}$ varies across subjects.
        In the case of categorical outcomes (such as \textsf{Return-6M}), chance is computed as 1 divided by the number of possible outcomes.
        
        The baseline model outputs the prior distribution of the target variables, $f\left(\boldsymbol{x}_{\textsf{out}}\right)$, which is computed from the corresponding marginal distribution of the $k=1$ model.
        This is equivalent to using the model with 1 latent component, as in this version all variables are assumed to be independent of each other, and consequently the input data have no impact on inference.
        This distribution is then used in (\ref{eq_EAE}) to compute the EAE.
        Performance of the baseline model is expected to be improved over chance.
        Comparing performance of models with order $>$ 1 to the baseline for each outcome variable enables us to measure the decrease in uncertainty, or equivalently the decrease in EAE that occurs when utilizing input data.
        
        \subsubsection{Population-based extrapolation risk evaluation} \label{sec_conf}
        Using the inference procedure, the model will output a distribution of unknown values given input data, $f\left(\boldsymbol{x}_{\textsf{out}}|\boldsymbol{x}_{\textsf{in}}\right)$.
        A question of interest is to determine ability of the model to accurately capture the likelihood in regions of the input space that are distinct from those covered by the training set.
        Since we cannot expect any single training set to cover all possible cases, we expect that our model may be less accurate for underrepresented patient populations.

        A reasonable approach to evaluate this risk is to use the trained model and consider the model fit of the input variables $f\left(\boldsymbol{x}_{\textsf{in}}\right)$ for an individual test subject.
        If this is high relative to the training population, then we may be confident that the model has been exposed to many similar samples in the training set.
        If, on the other hand, the likelihood of the input variables is low relative to the population, then the implication is that the model has been trained on few similar examples during training.
        It is possible that this value may frequently be low, since there are a large number of variables compared to the number of subjects and this makes it easy to come up with a combination of values that is not similar to any patients in the training set.
        
        The input likelihood
        is computed by first evaluating the marginal likelihood of the input data alone, using the marginal product distributions for each mixture component of the model (\ref{eq_model}),
        \begin{equation*}
            f\left(\boldsymbol{x}_{\textsf{in}};\boldsymbol{\theta}_z\right) = \prod_{v\in V_{\textsf{in}}} f\left(x_v ; \boldsymbol{\theta}_{z,v}\right),
        \end{equation*}
        where $V_{\textsf{in}}$ is the set of input variable indices.
        The input test likelihood
        is computed by,
        \begin{equation}
            \label{eq_conf}
            c = \sum_{z=1}^{k} \alpha_z
            f\left(\boldsymbol{x}_{\textsf{in}} ; \boldsymbol{\theta}_z\right).
        \end{equation}
        
\section{Results} \label{sec_results}

    \subsection{Model Selection and Unsupervised Learning}

\begin{figure}[!t]
    \centering
        \includegraphics[width=2.5in]{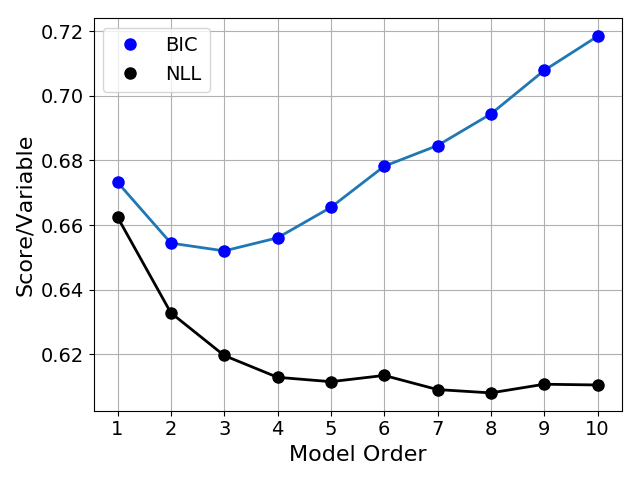}
        \caption{Bayesian Information Criterion (BIC) and Negative Log-Likelihood (NLL) curves for model orders 1-10. The NLL continues to decrease with increased model order, indicating improved model fit. However, the BIC is lowest for the model with order $k=3$.}
    \label{fig_bic}
    \end{figure}

    Fig. \ref{fig_bic} shows the BIC scores and the model fit (Negative Log-Likelihood, NLL) as a function of model order.
    The model was trained on all data, including 3, 6, and 12 month outcome variables.
    While the NLL decreases and converges with increasing model order, the BIC reaches a minimum value at $k=3$.
    Note that this minimum value is a function of the data, and we expect a higher value with greater sample sizes.
    
\begin{figure*}[!t]
    \centering
        \subfloat[\textsf{GOSE-6M}]{\includegraphics[width=2.5in]{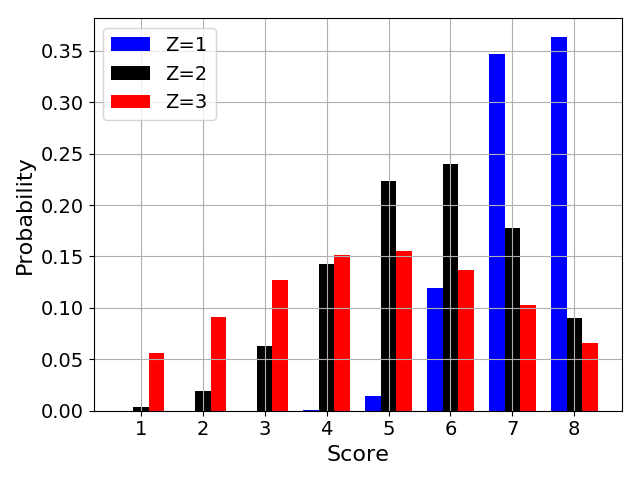}%
        \label{fig_gose6}}
        \hfil
        \subfloat[\textsf{PCL-6M}]{\includegraphics[width=2.5in]{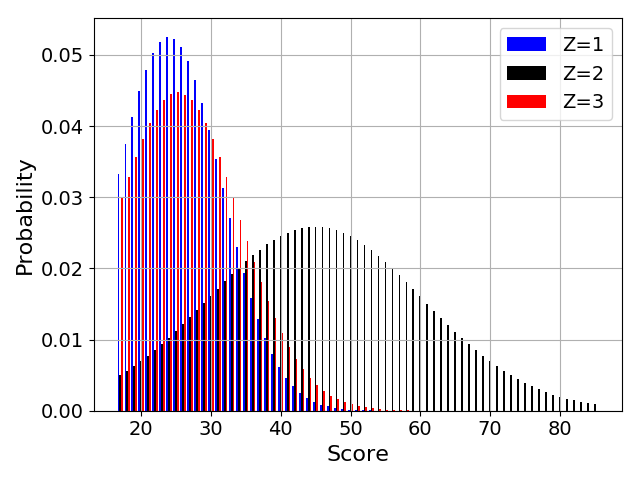}%
        \label{fig_pcl6}}
        \hfil
        \subfloat[\textsf{CT Marshall Score}]{\includegraphics[width=2.5in]{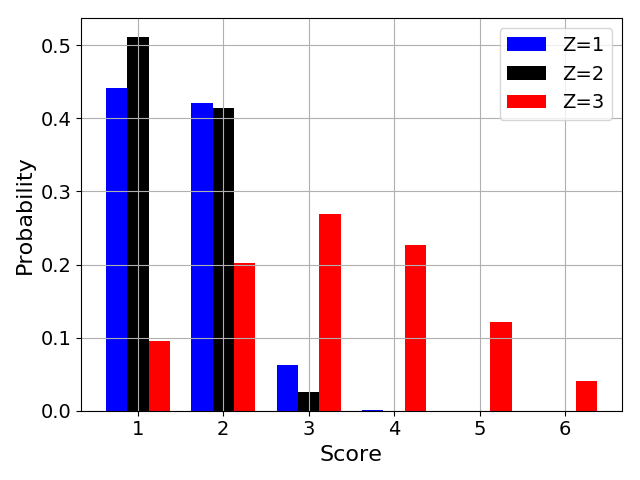}%
        \label{fig_ctmarshall}}
        \hfil
        \subfloat[\textsf{UCHL-1}]{\includegraphics[width=2.5in]{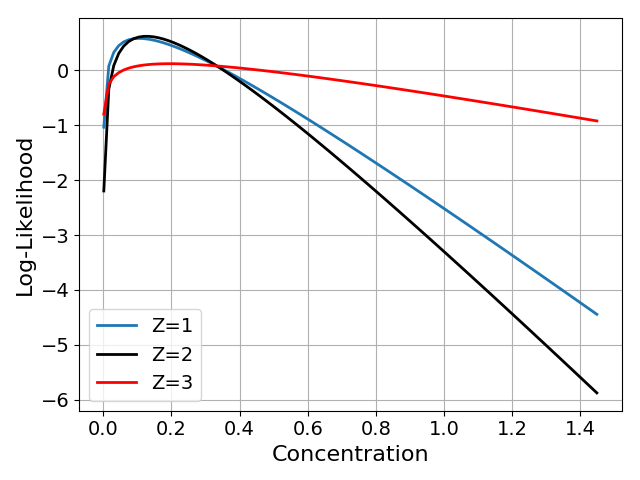}%
        \label{fig_uchl1}}
        \caption{Component distributions for selected variables using the model with the lowest BIC score, containing $k = 3$ mixture components: (a) \textsf{GOSE-6M}, (b) \textsf{PCL-6M}, (c) \textsf{Marshall} and (d) \textsf{UCH-L1}.}
        \label{fig_params}
    \end{figure*}

    Using the $k=3$ model, we take a closer look at the latent components for several selected variables: \textsf{GOSE-6M}, \textsf{PCL-6M}, \textsf{CT-Marshall}, and \textsf{UCHL-1}.
    We choose these variables to show variety in variable categories, containing outcomes, imaging annotations, and blood measures.
    Fig. \ref{fig_params} shows the 3 component distributions for these four variables.
    As can be seen from the figure, the three components for \textsf{GOSE-6M} cover broad categories of global recovery, ranging from best ($Z=1$) to worst ($Z=3$).
    The \textsf{PCL-6M} distributions show that the second component ($Z=2$) has the highest likelihood of severe PTSD symptoms, as measured by this variable.
    Greater likelihood for high \textsf{CT-Marshall} scores in the $Z=3$ component can be seen from the distributions in Fig. \ref{fig_ctmarshall}.
    The measure of \textsf{UCHL-1} concentration shows that high concentrations favor the more severe global outcome group, $Z=3$.
    To get a broader sense of these 3 components, the mode of these and several more selected variables for each component are shown in Table \ref{tbl_modes}.
    
    \begin{table}
        \renewcommand{\arraystretch}{1.3}
        \caption{Distribution Modes by Component}
        \label{tbl_modes}
        \centering
        \begin{tabular}{l||c|c|c}
        \textbf{Variable} & \textbf{C1} & \textbf{C2} & \textbf{C3}\\
        \hline
        \textsf{Age} &   41 & 44 & 47\\
        \textsf{GCS}	&   15 & 14 & 8\\
        \textsf{UCH-L1}  &   0.11    &   0.13    &   0.20\\
        \textsf{Marshall} &   1 & 1   &   3\\
        \textsf{GOSE-6M}	&   8 & 6 & 5\\
        \textsf{SWLS-6M}  &   25 & 16 & 24\\
        \textsf{PCL-6M} & 24 & 45 & 25\\
        \textsf{RPQ-6M} & 3 & 22 & 8\\
        \end{tabular}
    \end{table}
    
    \subsection{Outcome Inference}
    In this section, we show prediction performance for several selected outcome variables: \textsf{GOSE-3M, GOSE-6M, GOSE-12M,Neuro-6M, Return-6M,} and \textsf{PCL-6M}.
    The distributions of these variables are computed using the input variables, $\boldsymbol{x}_{\textsf{in}}$, which includes all of the non-outcome variables: demographics, prior medical history, clinical measurements, injury details, blood specimens, and imaging variables.
    We use the leave-one-out validation scheme described in Section \ref{sec_perf} and compare our results to those of the chance classifier and baseline models (Section \ref{sec_chancebase}).
    
    Table \ref{tbl_perf} shows prediction performance for the 6 outcome variables considered.
    Five of these variables (\textsf{GOSE-3M, GOSE-6M, GOSE-12M, Neuro-6M, and PCL-6M}) are ordinal valued, and for these we compute the expected absolute error, as defined in Section \ref{sec_perf}.
    For the \textsf{Return-6M} variable, which is a categorical variable, we report the probability of error.
    The values reported in Table \ref{tbl_perf} are normalized so that 100\% is the highest possible EAE.
    This is done by computing,
    \begin{equation} \label{eq_EAE_norm}
        \text{EAE}_{\text{norm}} = 100 \times \frac{\text{EAE}}{\text{max}(\text{EAE})},
    \end{equation}
    where $\text{max}(\text{EAE})$ is the maximum possible error.
    For Ordinal variables, this is the number of possible values reduced by 1 and for categorical variables this is equal to 1.
    Specifically, these maximum values are 7, 5, 1, and 68 for \textsf{GOSE, Neuro, Return,} and \textsf{PCL} variables, respectively.
    
    The column headings in Table \ref{tbl_perf} correspond to the model order, with $k=0$ signifying the uniformly random chance classifier.
    The $k=1$ column is the baseline classifier that utilizes prior distributions of the outcomes only.

    Values in parenthesis represent 2 standard deviations as measured through the leave-one-out procedure.
    The lowest normalized EAE is highlighted for each variable.
    Note that the optimal model in the BIC sense is not always the best performer, but the performance tends to be similar across model orders from $k=2$ and greater.
    
    \begin{table*}[!t]
    \renewcommand{\arraystretch}{1.3}
    \caption{Outcome Inference Performance. Normalized expected absolute error using (\ref{eq_EAE_norm}).}
    \label{tbl_perf}
    \centering
    \begin{tabular}{c||c|c|c|c|c|c|c}

Variable Name &   $k=0$ &   $k=1$	&   $k=2$ &   $k=3$ &   $k=4$ &   $k=5$ &   $k=6$\\

\hline

\textsf{GOSE-3M} &  39.71 (16.43)  &  27.71 (23.71)  &  \bf{21.71 (26.43)}  &  22.00 (24.57)  &  21.86 (24.14)  &  22.43 (23.57)  &  22.14 (24.00)\\
\textsf{GOSE-6M}	&  40.29 (16.57)  &  28.29 (25.00)  &  \bf{22.43 (30.00)}  &  22.57 (27.86)  &  23.29 (26.86)  &  23.86 (26.86)  &  23.43 (27.14)\\
\textsf{GOSE-12M}    &  41.86 (16.57)  &  25.00 (24.14)  &  \bf{20.43 (28.29)}  &  20.57 (26.43)  &  21.29 (24.86)  &  21.86 (26.29)  &  21.57 (25.71)\\
\textsf{Neuro-6M}   &  39.60 (17.00)  &  30.80 (22.60)  &  28.40 (34.80)  &  28.20 (30.20)  &  \bf{28.00 (30.00)}  &  28.80 (31.40)  &  28.40 (30.20)\\
\textsf{Return-6M}	&  80.00 (0.00)  &  65.00 (32.00)  &  \bf{59.00 (55.00)}  &  60.00 (50.00)  &  \bf{59.00 (53.00)}  &  61.00 (49.00)  &  60.00 (51.00)\\
\textsf{PCL-6M}   &  36.94 (16.69)  &  22.93 (16.93)  &  21.91 (29.29)  &  19.90 (23.40)  &  20.38 (24.66)  &  20.09 (24.65)  &  \bf{19.56 (23.72)}\\
\end{tabular}

    \end{table*}
    
\begin{figure}[!t]
    \centering
        \includegraphics[width=2.5in]{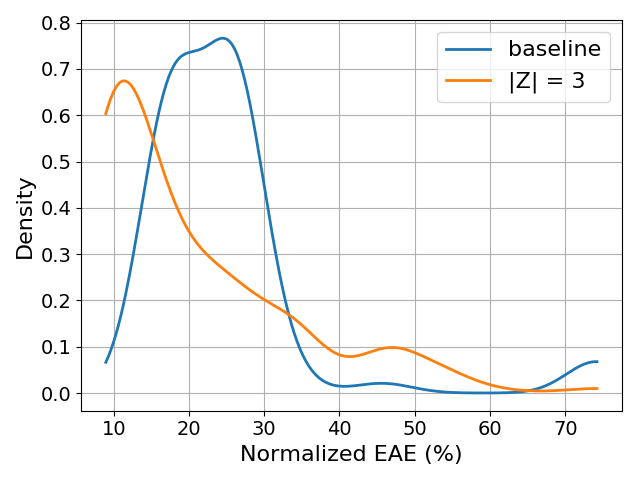}
        \caption{Distributions of the Expected Absolute Error (EAE) of the \textsf{GOSE-12M} outcome variable for the baseline and $k=3$ models. One EAE is computed for each subject using each model in a leave-one-out procedure.}
    \label{fig_EAE_dist}
    \end{figure}

    In addition to evaluating average performance, we are interested in examining individual variability in the expected error.
    Fig. \ref{fig_EAE_dist} shows  normalized EAE distributions across subjects for the \textsf{GOSE-12M} for the baseline and $k=3$ models.\footnote{This smoothed distribution was computed using a Gaussian Kernel Density Estimate using the bandwidth selection technique of Scott \cite{Scott2015-su}.}
    Note that the mode of the $k=3$ distribution is less than 15\%, whereas the normalized EAE across subjects for this model is 20.57\% (Table \ref{tbl_perf}).
    
    \subsection{Extrapolation Risk Evaluation}
    We are motivated by the variability in performance across subjects (as seen in Figure \ref{fig_EAE_dist}) to evaluate individualized model performance.
    By computing the model fit on the input data of a test subject, we are measuring how similar this subject's input data is to the training population.
    This score (Section \ref{sec_conf}) is based on the likelihood of the input data $\boldsymbol{x}_{\textsf{in}}$ given the model (\ref{eq_conf}).
    Since only the input data are used, this approach can be used with patients whose outcomes have not yet been observed.
    
    A normalization of the likelihoods can be used so that we can compare
    likelihoods
    computed from different models trained during the leave-one-out process and objectively interpret their meaning.
    In order to accomplish this, we compute the 
    input test likelihood
    for all subjects that the model was trained on, $\boldsymbol{c}_{\textsf{train}}$.
    Then we compute the percentile ranking of this score for test subject, $c$,
    $$
        p = \frac{\sum_{c' \in \boldsymbol{c}_{\textsf{train}}} I_{c>c'}}{|\boldsymbol{c}_{\textsf{train}}|},
    $$
    where $|\boldsymbol{c}_{\textsf{train}}|$ is the number of training subjects.
    This quantity is dependent on the training set and we use it to compare across models generated in the leave-one-out process, since the percentile is a normalized quantity.
    
\begin{figure}[!t]
    \centering
        \includegraphics[width=2.5in]{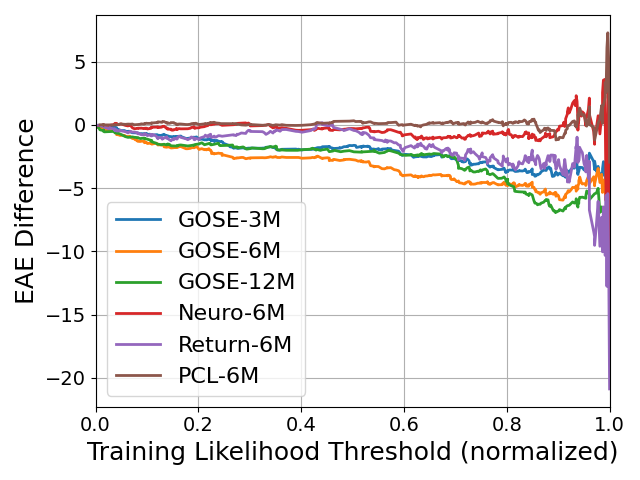}
        \caption{Cumulative expected error as a function of threshold. For a given threshold, the expected errors of all subjects with input test likelihood greater than this threshold are shown for six outcome measures.}
    \label{fig_conf_thresh}
    \end{figure}

    Fig. \ref{fig_conf_thresh} shows how the normalized EAE changes as subjects with percentile ranked input likelihood scores below a threshold are removed.
    The EAE is computed for all subjects with percentile ranked input likelihood greater than the threshold $\tau$,
    $$
        E\left(\tau\right)=\frac{1}{N_\tau} \sum_s I_{(p_s \ge \tau)} \text{EAE}_s,
    $$
    where $N_\tau$ is the number of subjects with likelihood greater than the threshold, $p_s$ is the percentile for subject $s$, and $\text{EAE}_s$ is the EAE for subject $s$.
    In Fig. \ref{fig_conf_thresh}, the difference $E\left(0\right) - E\left(\tau\right)$ is shown.
    A threshold of $\tau=0$ corresponds to all subjects being included in the computation of the EAE.
    As the threshold is increased, fewer subjects are included and this consequentially results in a noisier portion of the curve.
    From the figure, we can see that for four of the outcome variables the general trend is downward, indicating that for these variables, higher input likelihood is generally associated with lower error.
    
    To evaluate low and high regions of risk, we choose to use a cutoff value of $\tau=0.5$ and compute EAE for subjects in the two resulting bins: input likelihood lower than 0.5 and greater than or equal to 0.5.
    Table \ref{tbl_conf_lowhigh} shows the normalized EAE for the same set of variables that are shown in Table \ref{tbl_perf} using these two bins.
    This technique reveals that those subjects in the higher likelihood bin (more similar to the training set) have lower EAE than those in the lower bin for most of these variables.
    
    \begin{table}
        \renewcommand{\arraystretch}{1.3}
        \caption{Expected Error By Input Likelihood Bin}
        \label{tbl_conf_lowhigh}
        \centering
        \begin{tabular}{l||c|c|c}
        \textbf{Outcome} & \textbf{$<$0.5 (\%)} & \textbf{$\ge$0.5 (\%)} & \textbf{Decrease (\%)}\\
        \hline
        \textsf{GOSE-3M} &   24.0 & 20.4 & 3.5\\
        \textsf{GOSE-6M}	&   26.2 & 19.7 & 6.5\\
        \textsf{GOSE-12M}  &   23.8    &   18.4    &   5.4\\
        \textsf{Neuro-6M} &   28.6 & 27.8   &   0.8\\
        \textsf{Return-6M}	&   60.4 & 59.3 & 1.2\\
        \textsf{PCL-6M} & 16.3 & 17.2 & -0.9\\
        \end{tabular}
    \end{table}

\section{Discussion} \label{sec_discussion}
The model interpretation shown for selected variables in Fig. \ref{fig_params} gives a picture of patient stratification.
Component 1 corresponds to good global function as measured by the GOSE and low likelihood of PTSD as measured by the PCL.
Component 2 has worse global function than component 1, but better outcomes than component 3, with low probability of death (GOSE=1).
However, despite intermediate global/functional outcome, component 2 has the most severe PTSD symptoms (Fig. \ref{fig_pcl6}).
This indicates that our model has differentiated between global disability level and PTSD symptoms.

The CT Marshall score distributions in Fig. \ref{fig_ctmarshall} show that component 3 is more likely to have worse CT characteristics.
Concentration of \textsf{UCH-L1} is also more likely to be higher than 0.5 pg/ml for component 3.

The component distribution modes in Table \ref{tbl_modes} help to further interpret the three components.
These results compliment those in Fig. \ref{fig_params}, with the addition of \textsf{Age}, \textsf{GCS}, \textsf{SWLS-6M}, and \textsf{RPQ-6M}.
The \textsf{Age} has a slight increase in mode as we move from component 1 to 3.
The \textsf{GCS} is significantly worse in component 3, which is consistent with the general trend of poor \textsf{GOSE-6M} outcomes in this component.
Component 2 seems to represent those patients who have better \textsf{GOSE-6M} outcomes compared to component 3, but have greater likelihood for TBI-related symptoms, including anxiety and depression.
This can be seen by the worse scores within the \textsf{PCL-6M} and \textsf{RPQ-6M} measures in Table \ref{tbl_modes}.
In addition, \textsf{SWLS-6M} is also worse for component 2.

The baseline method does not use any information from the input data to make predictions, and only corrects for the base rate of outcome scores in the population.
Equivalently, this baseline could be viewed as the best guess if only population statistics about the outcomes were known and not any individualized information.
Our inference results (Table \ref{tbl_perf}) show that the model is capturing information in the input data that are predictive of outcomes, and therefore reduces uncertainty in the outcomes given the inputs.
We note that the baseline performance is improved over chance for all variables reported, implying that there is information contained in population statistics of the outcomes to perform better than chance without having access to the input data.

We chose the six outcome variables in Table \ref{tbl_perf} as they are commonly used and represent a variety of measures, including global function (GOSE), neurological (Neuro), anxiety/PTSD (PCL), and social status (Return) outcomes.
While we have not evaluated the performance of all outcome variables, a selection of others that we tested showed that in most cases the model errors are lower than the baseline.
In some cases the results are similar to the baseline, and in one instance (for a measure of functional independence, FIM) the model errors were larger than the baseline.
In this latter case (the \textsf{FIM-6M}), we hypothesize that the result is due to a small number of samples available and a preponderance of skewed scores.
A full analysis of the entire scope of all 352 outcome measures is possible with this approach and is left for future work.

Note that chance prediction ($k=0$) has variability for ordinal variables (\textsf{GOSE-3M, GOSE-6M, GOSE-12M, Neuro-6M, PCL-6M} in Table \ref{tbl_perf}) since the EAE (\ref{eq_EAE}) is evaluated empirically for each held-out subject.
True values in the middle of possible ordinal valued outcomes will have lower EAE than low or high values, which accumulate higher errors scored against the uniform distribution.
Categorical variables, such as \textsf{Return-6M}, do not vary for chance prediction since the probability of error is always $1-1/V$, where $V$ is the number of possible values.

From the model selection results (Fig. \ref{fig_bic}), we can see that the model with $k=3$ is optimal with respect to the BIC.
However, this does not imply that the $k=3$ optimal model will have the best prediction performance.
From Table \ref{tbl_perf}, we can see that the errors are relatively constant as the model order increases from 2 to 6.
The exception is \textsf{PCL-6M}, which shows a mean decrease from 18.62 to 16.91 as the model order increases from 2 to 3.
One possible explanation for this consistency in prediction as a function of model order is that higher order models may be attributing the additional components to small sub-populations of subjects which do not affect the EAE greatly.
This could leave only a few components that capture most of the variability in the data and that are similar across models.

Fig. \ref{fig_EAE_dist} shows there is variability in predication performance across subjects for the \textsf{GOSE-12M}.
The baseline approach is worse on average (25.00 vs. 20.43 normalized EAE), however some subjects perform better in the baseline version.
This is the case because there are some subjects with very high errors ($>40$ normalized EAE) in the $k=3$ model, while the likelihood of such large errors is low in the baseline.
This greater variability in the EAE may be due to erroneous associations that are captured in the model, but do not appear in all subjects.

Based on this analysis, one question of clinical value is to be able to determine whether or not the model's inference of outcomes should be trusted for a specific individual.
One approach for evaluating this level of extrapolation risk is presented in this paper.
Our results show that there is an association between the input test likelihood and prediction error (Fig. \ref{fig_conf_thresh} and Table \ref{tbl_conf_lowhigh}).
As can be seen in this figure, there is a downward trend in the expected error as the threshold increases for several variables.
Performance is relatively constant for the \textsf{PCL-6M} and \textsf{Neuro-6M} as a function of threshold.
The results in Table \ref{tbl_conf_lowhigh} shows performance separately for two equally spaced bins.
As can be seen in the table, the results for some of the variables (\textsf{Neuro-6M}, \textsf{Return-6M}, and \textsf{PCL-6M}) show little difference between the two bins.
We hypothesize that this may be caused the high-dimensionality of the input space and may by alleviated with a larger number of subjects in the training set.


We offer a few differentiating aspects of our approach to purely predictive ones.
First, our approach allows for prediction of many outcome metrics simultaneously with a single model, using any subset of input variables, avoiding the need to retrain a predictor for each outcome metric.
This is particularly important in large-scale studies such as TRACK Pilot, where a large outcome space consisting of several hundred variables exists.
In addition, training a large number of separate predictive models may result in spurious positive results, requiring careful validation.
Another possible advantage is in our handling of missing values,
where we do not impute missing values and therefore avoid the problem of choosing an imputation method.
Since we explicitly model missingness, we do not need to transform the data or impute missing values, which may introduce artifacts and additional biases.
If required for other methods, we can impute missing values using a trained model, either by selecting the modes of the resulting distributions of missing values, or performing multiple imputation by sampling from these distributions.

However, the development of predictive models is also an important line of work.
Since our framework is not optimized for prediction, it may be possible to outperform our prediction results using finely tuned predictive machine learning methods.
Some variables may be more amenable to purely prediction approaches, particularly non-clinically determined outcome variables such as the return to work status (\textsf{Return}).
It would benefit from the comparison of multiple classifiers, such as the decision tree, support vector machine, and multi-layer perceptron, and require the careful validation of hyperparameter selection.
The problem of input variable selection is also important in such predictive methods since the number of input variables is large relative to the number of subjects.

The problem of scoring the relevance of input variables toward each outcome is of high interest.
There exist many different feature scoring methods that can be applied to this problem.
In addition, the mixture model itself can be leveraged for this purpose as the full joint distribution between inputs and outcomes is specified.
This aspect of the problem is left as future work.

\section{Conclusion} \label{sec_conclusion}
The aim of this work is the development of a generative probabilistic data model than can be employed in TBI informatics to aid in clinical prognosis.
For example, this work revealed 3 broad groups of TBI patients with clinically distinct injury features and outcomes.
Furthermore, conditional distributions derived could be used to make diverse prognostic estimates across the entire dataset (e.g., likelihood of good functional outcome from age and injury parameters).
Our approach is broad enough to be used in several different ways, including patient phenotyping, patient grouping/stratification, outcome inference, and random sampling.
One advantage of our approach is that we are able to infer all of the outcome variables simultaneously with a single model.
Using a standard ML approach would require separate predictors for each outcome, raising the increased possibility of spurious results.
However, this is an area of research for future work.


The input likelihood scoring technique addresses the important question of model reliability.
The approach described in this work is based on the notion that a prediction with a model having a higher model fit should be trusted more than one with a lower fit.
Effectively, this is an evaluation of the similarity of a test subject with the training population.
We are motivated by the idea that prognosis of a test subject that is not similar to the training population may not be reliable.

Our method can be used for both prediction and unsupervised learning.
The component distributions are important for interpretation of the results and patient stratification.
Interpretation of the predictive results with respect to these latent component training in an unsupervised manner could yield useful insights.
As TBI is complex in both the variety of outcomes and changes over time, uncovering structure in the data which can be interpreted against existing clinical knowledge is an important goal of data driven research in this field.

The EAE metric used is computed using the inferred output distribution for the target outcome variable.
In contrast to predicting a single value for the outcome, we instead consider all possible values that could be predicted and weigh them by their inferred likelihoods.
It is also possible to instead compute the absolute error by choosing the single most likely value for each prediction.
In this version, a hard decision for each subject would be made and the error from this single predicted value computed, resulting in the Mean Absolute Error (MAE) evaluation.

Within this framework it is possible to incorporate additional data elements as they become available.
In addition, as sample sizes increase we expect the model to produce more accurate distributions and prediction results.
One aspect of future work is the evaluation of the most relevant input variables towards a prediction.
We decided to keep as many variables as possible in our analysis, but it may be the case that they are not all required to make accurate predictions.
Several approaches could be considered for this, including ablation of inputs or conditional entropy calculations of individual input variables and outcomes.
We also note that as sample sizes increase the set of variables which may have been deemed to be more informative can change.
In this sense, a framework which is capable of incorporating all of the available data is desirable.

\section*{Acknowledgment}
This work was performed under the auspices of the U.S. Department of Energy by Lawrence Livermore National Laboratory under Contract DE-AC52-07NA27344.

We would like to thank Kristofer Bouchard and Andrew Tritt from Lawrence Berkeley National Laboratory and Neil Getty from Argonne National Laboratory for conversations related to modeling TRACK-Pilot data.
We would like to thank Grant Bouquet from Lawrence Livermore National Laboratory for conversations on software tools for probabilistic modeling.

\ifCLASSOPTIONcaptionsoff
  \newpage
\fi

\bibliography{tbi_refs}{}

\begin{thebibliography}{10}
\providecommand{\url}[1]{#1}
\csname url@samestyle\endcsname
\providecommand{\newblock}{\relax}
\providecommand{\bibinfo}[2]{#2}
\providecommand{\BIBentrySTDinterwordspacing}{\spaceskip=0pt\relax}
\providecommand{\BIBentryALTinterwordstretchfactor}{4}
\providecommand{\BIBentryALTinterwordspacing}{\spaceskip=\fontdimen2\font plus
\BIBentryALTinterwordstretchfactor\fontdimen3\font minus
  \fontdimen4\font\relax}
\providecommand{\BIBforeignlanguage}[2]{{%
\expandafter\ifx\csname l@#1\endcsname\relax
\typeout{** WARNING: IEEEtran.bst: No hyphenation pattern has been}%
\typeout{** loaded for the language `#1'. Using the pattern for}%
\typeout{** the default language instead.}%
\else
\language=\csname l@#1\endcsname
\fi
#2}}
\providecommand{\BIBdecl}{\relax}
\BIBdecl

\bibitem{Maas2017-yb}
A.~I.~R. Maas, D.~K. Menon, P.~D. Adelson, N.~Andelic, M.~J. Bell, A.~Belli,
  P.~Bragge, A.~Brazinova, A.~B{\"u}ki, R.~M. Chesnut, G.~Citerio, M.~Coburn,
  D.~J. Cooper, A.~T. Crowder, E.~Czeiter, M.~Czosnyka, R.~Diaz-Arrastia, J.~P.
  Dreier, A.-C. Duhaime, A.~Ercole, T.~A. van Essen, V.~L. Feigin, G.~Gao,
  J.~Giacino, L.~E. Gonzalez-Lara, R.~L. Gruen, D.~Gupta, J.~A. Hartings,
  S.~Hill, J.-Y. Jiang, N.~Ketharanathan, E.~J.~O. Kompanje, L.~Lanyon,
  S.~Laureys, F.~Lecky, H.~Levin, H.~F. Lingsma, M.~Maegele, M.~Majdan,
  G.~Manley, J.~Marsteller, L.~Mascia, C.~McFadyen, S.~Mondello, V.~Newcombe,
  A.~Palotie, P.~M. Parizel, W.~Peul, J.~Piercy, S.~Polinder, L.~Puybasset,
  T.~E. Rasmussen, R.~Rossaint, P.~Smielewski, J.~S{\"o}derberg, S.~J.
  Stanworth, M.~B. Stein, N.~von Steinb{\"u}chel, W.~Stewart, E.~W. Steyerberg,
  N.~Stocchetti, A.~Synnot, B.~Te~Ao, O.~Tenovuo, A.~Theadom, D.~Tibboel,
  W.~Videtta, K.~K.~W. Wang, W.~H. Williams, L.~Wilson, K.~Yaffe, and {InTBIR
  Participants and Investigators}, ``\BIBforeignlanguage{en}{Traumatic brain
  injury: integrated approaches to improve prevention, clinical care, and
  research},'' \emph{\BIBforeignlanguage{en}{Lancet Neurol.}}, vol.~16, no.~12,
  pp. 987--1048, Dec. 2017.

\bibitem{Dewan2018-vh}
M.~C. Dewan, A.~Rattani, S.~Gupta, R.~E. Baticulon, Y.-C. Hung, M.~Punchak,
  A.~Agrawal, A.~O. Adeleye, M.~G. Shrime, A.~M. Rubiano, J.~V. Rosenfeld, and
  K.~B. Park, ``\BIBforeignlanguage{en}{Estimating the global incidence of
  traumatic brain injury},'' \emph{\BIBforeignlanguage{en}{J. Neurosurg.}}, pp.
  1--18, Apr. 2018.

\bibitem{Millis2001-wh}
S.~R. Millis, M.~Rosenthal, T.~A. Novack, M.~Sherer, T.~G. Nick, J.~S.
  Kreutzer, W.~M. High, Jr, and J.~H. Ricker,
  ``\BIBforeignlanguage{en}{Long-term neuropsychological outcome after
  traumatic brain injury},'' \emph{\BIBforeignlanguage{en}{J. Head Trauma
  Rehabil.}}, vol.~16, no.~4, pp. 343--355, Aug. 2001.

\bibitem{Rabinowitz2014-hc}
A.~R. Rabinowitz and H.~S. Levin, ``\BIBforeignlanguage{en}{Cognitive sequelae
  of traumatic brain injury},'' \emph{\BIBforeignlanguage{en}{Psychiatr. Clin.
  North Am.}}, vol.~37, no.~1, pp. 1--11, Mar. 2014.

\bibitem{Dikmen2017-xe}
S.~Dikmen, J.~Machamer, and N.~Temkin, ``\BIBforeignlanguage{en}{Mild traumatic
  brain injury: Longitudinal study of cognition, functional status, and
  {Post-Traumatic} symptoms},'' \emph{\BIBforeignlanguage{en}{J. Neurotrauma}},
  vol.~34, no.~8, pp. 1524--1530, Apr. 2017.

\bibitem{Mollayeva2019-gl}
T.~Mollayeva, S.~Mollayeva, N.~Pacheco, A.~D'Souza, and A.~Colantonio,
  ``\BIBforeignlanguage{en}{The course and prognostic factors of cognitive
  outcomes after traumatic brain injury: A systematic review and
  meta-analysis},'' \emph{\BIBforeignlanguage{en}{Neurosci. Biobehav. Rev.}},
  vol.~99, pp. 198--250, Apr. 2019.

\bibitem{Maas2010-qc}
A.~I. Maas, C.~L. Harrison-Felix, D.~Menon, P.~D. Adelson, T.~Balkin,
  R.~Bullock, D.~C. Engel, W.~Gordon, J.~L. Orman, H.~L. Lew, C.~Robertson,
  N.~Temkin, A.~Valadka, M.~Verfaellie, M.~Wainwright, D.~W. Wright, and
  K.~Schwab, ``\BIBforeignlanguage{en}{Common data elements for traumatic brain
  injury: recommendations from the interagency working group on demographics
  and clinical assessment},'' \emph{\BIBforeignlanguage{en}{Arch. Phys. Med.
  Rehabil.}}, vol.~91, no.~11, pp. 1641--1649, Nov. 2010.

\bibitem{Yue2013-gz}
J.~K. Yue, M.~J. Vassar, H.~F. Lingsma, S.~R. Cooper, D.~O. Okonkwo, A.~B.
  Valadka, W.~A. Gordon, A.~I.~R. Maas, P.~Mukherjee, E.~L. Yuh, A.~M. Puccio,
  D.~M. Schnyer, G.~T. Manley, and {TRACK-TBI Investigators},
  ``\BIBforeignlanguage{en}{Transforming research and clinical knowledge in
  traumatic brain injury pilot: multicenter implementation of the common data
  elements for traumatic brain injury},'' \emph{\BIBforeignlanguage{en}{J.
  Neurotrauma}}, vol.~30, no.~22, pp. 1831--1844, Nov. 2013.

\bibitem{Steyerberg2008-tv}
E.~W. Steyerberg, N.~Mushkudiani, P.~Perel, I.~Butcher, J.~Lu, G.~S. McHugh,
  G.~D. Murray, A.~Marmarou, I.~Roberts, J.~D.~F. Habbema, and A.~I.~R. Maas,
  ``\BIBforeignlanguage{en}{Predicting outcome after traumatic brain injury:
  development and international validation of prognostic scores based on
  admission characteristics},'' \emph{\BIBforeignlanguage{en}{PLoS Med.}},
  vol.~5, no.~8, p. e165; discussion e165, Aug. 2008.

\bibitem{Roozenbeek2012-fg}
B.~Roozenbeek, H.~F. Lingsma, F.~E. Lecky, J.~Lu, J.~Weir, I.~Butcher, G.~S.
  McHugh, G.~D. Murray, P.~Perel, A.~I. Maas, E.~W. Steyerberg, {International
  Mission on Prognosis Analysis of Clinical Trials in Traumatic Brain Injury
  (IMPACT) Study Group}, {Corticosteroid Randomisation After Significant Head
  Injury (CRASH) Trial Collaborators}, and {Trauma Audit and Research Network
  (TARN)}, ``\BIBforeignlanguage{en}{Prediction of outcome after moderate and
  severe traumatic brain injury: external validation of the international
  mission on prognosis and analysis of clinical trials ({IMPACT}) and corticoid
  randomisation after significant head injury ({CRASH}) prognostic models},''
  \emph{\BIBforeignlanguage{en}{Crit. Care Med.}}, vol.~40, no.~5, pp.
  1609--1617, May 2012.

\bibitem{Silverberg2015-xo}
N.~D. Silverberg, A.~J. Gardner, J.~R. Brubacher, W.~J. Panenka, J.~J. Li, and
  G.~L. Iverson, ``\BIBforeignlanguage{en}{Systematic review of multivariable
  prognostic models for mild traumatic brain injury},''
  \emph{\BIBforeignlanguage{en}{J. Neurotrauma}}, vol.~32, no.~8, pp. 517--526,
  Apr. 2015.

\bibitem{Shaker2015-yo}
M.~Shaker, D.~Erdogmus, J.~Dy, and S.~Bouix, ``Sparse model learning for high
  dimensional diffusion {MRI} data in traumatic brain injury,'' in \emph{2015
  {IEEE} 25th International Workshop on Machine Learning for Signal Processing
  ({MLSP})}, Sep. 2015, pp. 1--6.

\bibitem{Mitra2016-vt}
J.~Mitra, K.-K. Shen, S.~Ghose, P.~Bourgeat, J.~Fripp, O.~Salvado, K.~Pannek,
  D.~J. Taylor, J.~L. Mathias, and S.~Rose,
  ``\BIBforeignlanguage{en}{Statistical machine learning to identify traumatic
  brain injury ({TBI}) from structural disconnections of white matter
  networks},'' \emph{\BIBforeignlanguage{en}{Neuroimage}}, vol. 129, pp.
  247--259, Apr. 2016.

\bibitem{Vergara2017-aw}
V.~M. Vergara, A.~R. Mayer, E.~Damaraju, K.~A. Kiehl, and V.~Calhoun,
  ``Detection of mild traumatic brain injury by machine learning classification
  using resting state functional network connectivity and fractional
  anisotropy,'' \emph{J. Neurotrauma}, vol.~34, no.~5, pp. 1045--1053, Mar.
  2017.

\bibitem{Taslimitehrani2014-cv}
V.~Taslimitehrani and G.~Dong, ``A new {CPXR} based logistic regression method
  and clinical prognostic modeling results using the method on traumatic brain
  injury,'' in \emph{2014 {IEEE} International Conference on Bioinformatics and
  Bioengineering}, Nov. 2014, pp. 283--290.

\bibitem{Nielson2017-nq}
J.~L. Nielson, S.~R. Cooper, J.~K. Yue, M.~D. Sorani, T.~Inoue, E.~L. Yuh,
  P.~Mukherjee, T.~C. Petrossian, J.~Paquette, P.~Y. Lum, G.~E. Carlsson, M.~J.
  Vassar, H.~F. Lingsma, W.~A. Gordon, A.~B. Valadka, D.~O. Okonkwo, G.~T.
  Manley, A.~R. Ferguson, and {TRACK-TBI Investigators}, ``Uncovering precision
  phenotype-biomarker associations in traumatic brain injury using topological
  data analysis,'' \emph{PLoS One}, vol.~12, no.~3, p. e0169490, Mar. 2017.

\bibitem{Yuh2014-wk}
E.~L. Yuh, S.~R. Cooper, P.~Mukherjee, J.~K. Yue, H.~F. Lingsma, W.~A. Gordon,
  A.~B. Valadka, D.~O. Okonkwo, D.~M. Schnyer, M.~J. Vassar, A.~I.~R. Maas,
  G.~T. Manley, and {TRACK-TBI INVESTIGATORS},
  ``\BIBforeignlanguage{en}{Diffusion tensor imaging for outcome prediction in
  mild traumatic brain injury: a {TRACK-TBI} study},''
  \emph{\BIBforeignlanguage{en}{J. Neurotrauma}}, vol.~31, no.~17, pp.
  1457--1477, Sep. 2014.

\bibitem{Huie2019-lq}
J.~R. Huie, R.~Diaz-Arrastia, J.~K. Yue, M.~D. Sorani, A.~M. Puccio, D.~O.
  Okonkwo, G.~T. Manley, A.~R. Ferguson, and T.-T. Investigators, ``Testing a
  multivariate proteomic panel for traumatic brain injury biomarker discovery:
  a {TRACK-TBI} pilot study,'' \emph{J. Neurotrauma}, vol.~36, no.~1, pp.
  100--110, 2019.

\bibitem{Stein2019-gi}
M.~B. Stein, S.~Jain, J.~T. Giacino, H.~Levin, S.~Dikmen, L.~D. Nelson, M.~J.
  Vassar, D.~O. Okonkwo, R.~Diaz-Arrastia, C.~S. Robertson, P.~Mukherjee,
  M.~McCrea, C.~L. Mac~Donald, J.~K. Yue, E.~Yuh, X.~Sun, L.~Campbell-Sills,
  N.~Temkin, G.~T. Manley, {TRACK-TBI Investigators}, O.~Adeoye, N.~Badjatia,
  K.~Boase, Y.~Bodien, M.~R. Bullock, R.~Chesnut, J.~D. Corrigan, K.~Crawford,
  R.~Diaz-Arrastia, S.~Dikmen, A.-C. Duhaime, R.~Ellenbogen, V.~R. Feeser,
  A.~Ferguson, B.~Foreman, R.~Gardner, E.~Gaudette, J.~T. Giacino, L.~Gonzalez,
  S.~Gopinath, R.~Gullapalli, J.~C. Hemphill, G.~Hotz, S.~Jain, F.~Korley,
  J.~Kramer, N.~Kreitzer, H.~Levin, C.~Lindsell, J.~Machamer, C.~Madden,
  A.~Martin, T.~McAllister, M.~McCrea, R.~Merchant, P.~Mukherjee, L.~D. Nelson,
  F.~Noel, D.~O. Okonkwo, E.~Palacios, D.~Perl, A.~Puccio, M.~Rabinowitz, C.~S.
  Robertson, J.~Rosand, A.~Sander, G.~Satris, D.~Schnyer, S.~Seabury,
  M.~Sherer, M.~B. Stein, S.~Taylor, A.~Toga, N.~Temkin, A.~Valadka, M.~J.
  Vassar, P.~Vespa, K.~Wang, J.~K. Yue, E.~Yuh, and R.~Zafonte,
  ``\BIBforeignlanguage{en}{Risk of posttraumatic stress disorder and major
  depression in civilian patients after mild traumatic brain injury: A
  {TRACK-TBI} study},'' \emph{\BIBforeignlanguage{en}{JAMA Psychiatry}},
  vol.~76, no.~3, pp. 249--258, Mar. 2019.

\bibitem{Wang2017-iz}
L.~Wang, ``Heterogeneous data and big data analytics,'' \emph{Automatic Control
  and Information Sciences}, vol.~3, no.~1, pp. 8--15, 2017.

\bibitem{Pavlidis2001-ds}
P.~Pavlidis, J.~Weston, J.~Cai, and W.~N. Grundy, ``Gene functional
  classification from heterogeneous data,'' in \emph{Proceedings of the fifth
  annual international conference on Computational biology}, ser. RECOMB
  '01.\hskip 1em plus 0.5em minus 0.4em\relax New York, NY, USA: Association
  for Computing Machinery, Apr. 2001, pp. 249--255.

\bibitem{Lewis2006-ay}
D.~P. Lewis, T.~Jebara, and W.~S. Noble, ``\BIBforeignlanguage{en}{Support
  vector machine learning from heterogeneous data: an empirical analysis using
  protein sequence and structure},''
  \emph{\BIBforeignlanguage{en}{Bioinformatics}}, vol.~22, no.~22, pp.
  2753--2760, Nov. 2006.

\bibitem{Xiang2018-ld}
L.~Xiang, G.~Zhao, Q.~Li, W.~Hao, and F.~Li, ``{TUMK-ELM}: A fast unsupervised
  heterogeneous data learning approach,'' \emph{IEEE Access}, vol.~6, pp.
  35\,305--35\,315, 2018.

\bibitem{Luo2018-mx}
Y.~Luo, Y.~Wen, and D.~Tao, ``\BIBforeignlanguage{en}{Heterogeneous multitask
  metric learning across multiple domains},''
  \emph{\BIBforeignlanguage{en}{IEEE Trans Neural Netw Learn Syst}}, vol.~29,
  no.~9, pp. 4051--4064, Sep. 2018.

\bibitem{McLachlan2004-mh}
G.~J. McLachlan and D.~Peel, \emph{\BIBforeignlanguage{en}{Finite Mixture
  Models}}.\hskip 1em plus 0.5em minus 0.4em\relax John Wiley \& Sons, Mar.
  2004.

\bibitem{Wei2020-qd}
Y.~Wei, Y.~Tang, and P.~D. McNicholas, ``\BIBforeignlanguage{en}{Flexible
  {High-Dimensional} unsupervised learning with missing data},''
  \emph{\BIBforeignlanguage{en}{IEEE Trans. Pattern Anal. Mach. Intell.}},
  vol.~42, no.~3, pp. 610--621, Mar. 2020.

\bibitem{Pivovarov2015-zu}
R.~Pivovarov, A.~J. Perotte, E.~Grave, J.~Angiolillo, C.~H. Wiggins, and
  N.~Elhadad, ``\BIBforeignlanguage{en}{Learning probabilistic phenotypes from
  heterogeneous {EHR} data},'' \emph{\BIBforeignlanguage{en}{J. Biomed.
  Inform.}}, vol.~58, pp. 156--165, Dec. 2015.

\bibitem{Mayhew2018-pt}
M.~B. Mayhew, B.~K. Petersen, A.~P. Sales, J.~D. Greene, V.~X. Liu, and T.~S.
  Wasson, ``\BIBforeignlanguage{en}{Flexible, cluster-based analysis of the
  electronic medical record of sepsis with composite mixture models},''
  \emph{\BIBforeignlanguage{en}{J. Biomed. Inform.}}, vol.~78, pp. 33--42, Feb.
  2018.

\bibitem{Nelson2017-hc}
L.~D. Nelson, J.~Ranson, A.~R. Ferguson, J.~Giacino, D.~O. Okonkwo, A.~Valadka,
  G.~Manley, and M.~McCrea, ``\BIBforeignlanguage{en}{Validating
  multidimensional outcome assessment using the {TBI} common data elements: An
  analysis of the {TRACK-TBI} pilot sample},'' \emph{\BIBforeignlanguage{en}{J.
  Neurotrauma}}, Jun. 2017.

\bibitem{Wilson1998-hs}
J.~T. Wilson, L.~E. Pettigrew, and G.~M. Teasdale,
  ``\BIBforeignlanguage{en}{Structured interviews for the glasgow outcome scale
  and the extended glasgow outcome scale: guidelines for their use},''
  \emph{\BIBforeignlanguage{en}{J. Neurotrauma}}, vol.~15, no.~8, pp. 573--585,
  Aug. 1998.

\bibitem{Teasdale1998-qp}
G.~M. Teasdale, L.~E. Pettigrew, J.~T. Wilson, G.~Murray, and B.~Jennett,
  ``\BIBforeignlanguage{en}{Analyzing outcome of treatment of severe head
  injury: a review and update on advancing the use of the glasgow outcome
  scale},'' \emph{\BIBforeignlanguage{en}{J. Neurotrauma}}, vol.~15, no.~8, pp.
  587--597, Aug. 1998.

\bibitem{Weathers1991-jt}
E.~W. Weathers, J.~A. Huska, and T.~M. Keane, ``The {PTSD} check-
  list--civilian version ({PCL-C}),'' National Center for PTSD, Boston Veterans
  Affairs Medical Center, 150 S. Huntington Avenue, Boston, MA 02130, Tech.
  Rep., 1991.

\bibitem{Diener1985-cu}
E.~Diener, R.~A. Emmons, R.~J. Larsen, and S.~Griffin,
  ``\BIBforeignlanguage{en}{The satisfaction with life scale},''
  \emph{\BIBforeignlanguage{en}{J. Pers. Assess.}}, vol.~49, no.~1, pp. 71--75,
  Feb. 1985.

\bibitem{King1995-uy}
N.~S. King, S.~Crawford, F.~J. Wenden, N.~E. Moss, and D.~T. Wade,
  ``\BIBforeignlanguage{en}{The rivermead post concussion symptoms
  questionnaire: a measure of symptoms commonly experienced after head injury
  and its reliability},'' \emph{\BIBforeignlanguage{en}{J. Neurol.}}, vol. 242,
  no.~9, pp. 587--592, Sep. 1995.

\bibitem{Teasdale1974-uu}
G.~Teasdale and B.~Jennett, ``\BIBforeignlanguage{en}{Assessment of coma and
  impaired consciousness. a practical scale},''
  \emph{\BIBforeignlanguage{en}{Lancet}}, vol.~2, no. 7872, pp. 81--84, Jul.
  1974.

\bibitem{Sternbach2000-xh}
G.~L. Sternbach, ``\BIBforeignlanguage{en}{The glasgow coma scale},''
  \emph{\BIBforeignlanguage{en}{J. Emerg. Med.}}, vol.~19, no.~1, pp. 67--71,
  Jul. 2000.

\bibitem{Marshall1992-pc}
L.~F. Marshall, S.~B. Marshall, M.~R. Klauber, M.~Van Berkum~Clark,
  H.~Eisenberg, J.~A. Jane, T.~G. Luerssen, A.~Marmarou, and M.~A. Foulkes,
  ``\BIBforeignlanguage{en}{The diagnosis of head injury requires a
  classification based on computed axial tomography},''
  \emph{\BIBforeignlanguage{en}{J. Neurotrauma}}, vol. 9 Suppl 1, pp. S287--92,
  Mar. 1992.

\bibitem{Day2010-ii}
I.~N.~M. Day and R.~J. Thompson, ``\BIBforeignlanguage{en}{{UCHL1} ({PGP} 9.5):
  neuronal biomarker and ubiquitin system protein},''
  \emph{\BIBforeignlanguage{en}{Prog. Neurobiol.}}, vol.~90, no.~3, pp.
  327--362, Mar. 2010.

\bibitem{Little2019-nn}
R.~J.~A. Little and D.~B. Rubin, \emph{\BIBforeignlanguage{en}{Statistical
  Analysis with Missing Data}}.\hskip 1em plus 0.5em minus 0.4em\relax John
  Wiley \& Sons, Apr. 2019.

\bibitem{Moon1996-kb}
T.~K. Moon, ``The expectation-maximization algorithm,'' \emph{IEEE Signal
  Process. Mag.}, vol.~13, no.~6, pp. 47--60, Nov. 1996.

\bibitem{Dempster1977-gh}
A.~P. Dempster, N.~M. Laird, and D.~B. Rubin, ``\BIBforeignlanguage{en}{Maximum
  likelihood from incomplete data via {theEMAlgorithm}},''
  \emph{\BIBforeignlanguage{en}{J. R. Stat. Soc.}}, vol.~39, no.~1, pp. 1--22,
  Sep. 1977.

\bibitem{Choi1969-ax}
S.~C. Choi and R.~Wette, ``Maximum likelihood estimation of the parameters of
  the gamma distribution and their bias,'' \emph{Technometrics}, vol.~11,
  no.~4, pp. 683--690, Nov. 1969.

\bibitem{Hastie2009-ba}
T.~Hastie, R.~Tibshirani, and J.~Friedman, \emph{\BIBforeignlanguage{en}{The
  Elements of Statistical Learning: Data Mining, Inference, and Prediction,
  Second Edition}}.\hskip 1em plus 0.5em minus 0.4em\relax Springer Science \&
  Business Media, Aug. 2009.

\bibitem{Scott2015-su}
D.~W. Scott, \emph{\BIBforeignlanguage{en}{Multivariate Density Estimation:
  Theory, Practice, and Visualization}}.\hskip 1em plus 0.5em minus 0.4em\relax
  John Wiley \& Sons, Mar. 2015.

\end{thebibliography}
\bibliographystyle{IEEEtran}

%








\end{document}